\documentclass[10pt,twocolumn,letterpaper]{article}

\usepackage{cvpr}              

\usepackage{graphicx}
\usepackage{amsmath}
\usepackage{amssymb}
\usepackage{booktabs}

\usepackage{amsfonts}
\usepackage{float}
\usepackage{array}
\usepackage{color}
\usepackage{tabularx}
\usepackage{longtable}
\usepackage{tabu}
\usepackage{multirow}
\usepackage{url}
\usepackage{bm}
\usepackage[noend]{algpseudocode}
\usepackage{algorithmicx,algorithm}
\usepackage{adjustbox}

\usepackage[accsupp]{axessibility}  

\definecolor{grey}{RGB}{130,130,130} 
\definecolor{black}{RGB}{0,0,0} 

\usepackage[pagebackref,breaklinks,colorlinks]{hyperref}

\usepackage[capitalize]{cleveref}
\crefname{section}{Sec.}{Secs.}
\Crefname{section}{Section}{Sections}
\Crefname{table}{Table}{Tables}
\crefname{table}{Tab.}{Tabs.}

\newenvironment{tightcenter}{%
  \setlength\topsep{3pt}
  \setlength\parskip{3pt}
  \begin{center}
  \begin{minipage}{.42\textwidth}
}{%
  \end{minipage}
  \end{center}
}

\def\cvprPaperID{7856} 
\def\confName{CVPR}
\def\confYear{2023}

\begin{document}

\title{Blind Image Quality Assessment via Vision-Language Correspondence: A Multitask Learning Perspective}

\author{Weixia Zhang$^{1}$, Guangtao Zhai$^{1}$, Ying Wei$^{2}$, Xiaokang Yang$^{1}$, Kede Ma$^{2,3}\thanks{Corresponding author.}$\\
$^1$ MoE Key Lab of Artificial Intelligence, AI Institute, Shanghai Jiao Tong University \\
$^2$ Department of Computer Science,  City University of Hong Kong\\
$^3$ Shenzhen Research Institute,  City University of Hong Kong\\
\texttt{\{zwx8981, zhaiguangtao, xkyang\}@sjtu.edu.cn}\\
\texttt{\{yingwei, kede.ma\}@cityu.edu.hk}}

\maketitle

\begin{abstract}
We aim at advancing blind image quality assessment (BIQA), which predicts the human perception of image quality without any reference information. We develop a general and automated multitask learning scheme for BIQA to exploit  auxiliary knowledge from other tasks, in a way that the model parameter sharing and the loss weighting are determined automatically. Specifically, we first describe all candidate label combinations (from multiple tasks) using a textual template, and compute the joint probability from the cosine similarities of the visual-textual embeddings. Predictions of each task can be inferred from the joint distribution, and optimized by carefully designed loss functions. Through comprehensive experiments on learning three tasks - BIQA, scene classification, and distortion type identification, we verify that the proposed BIQA method 1) benefits from the  scene classification and distortion type identification tasks and outperforms the state-of-the-art on multiple IQA datasets, 2) is  more robust in the group maximum differentiation competition, and 3) realigns the quality annotations from different IQA datasets more effectively. The source code is available at \url{https://github.com/zwx8981/LIQE}.
\end{abstract}

\section{Introduction}
\label{sec:intro}
As a fundamental computational vision task, blind image quality assessment (BIQA)~\cite{wang2006modern} aims to predict the visual quality of a digital image with no access to the underlying pristine-quality counterpart (if any).  In the age of deep learning, the development of BIQA can be mainly characterized by strategies to mitigate the conflict between the large number of trainable parameters  and the small number of human quality annotations in the form of mean opinion scores (MOSs). When synthetic distortions (\eg, Gaussian noise and JPEG compression) are of primary concern, patchwise training~\cite{bosse2016deep}, quality-aware pre-training~\cite{liu2017rankiqa, Ma2018End, zhang2020blind}, and learning from noisy pseudo-labels~\cite{ma2019blind, athar2019comprehensive, wu2020end} are practical training tricks with less (or no) reliance on MOSs. Here the underlying assumptions are that 1) the pristine-quality images exist and are accessible, 2) the visual distortions can be simulated efficiently and automatically, and 3) full-reference IQA models~\cite{wang2004image} are applicable and provide adequate quality approximations. However, all these assumptions do not hold when it comes to realistic camera distortion (\eg, sensor noise, motion blurring or a combination of both). A different set of training tricks have been explored, including transfer learning~\cite{zhang2020blind,hosu2020koniq}, 
meta learning~\cite{zhu2020metaiqa}, and contrastive learning~\cite{madhusudana2022image}. Emerging techniques that combine multiple datasets for joint training~\cite{zhang2021uncertainty} and  that identify informative samples for active fine-tuning~\cite{Wang2021troubleshooting} can also be seen as ways to confront the data challenge in BIQA.

\begin{figure}[t]
    \centering
    \subfloat[]{\includegraphics[width=0.15\textwidth]{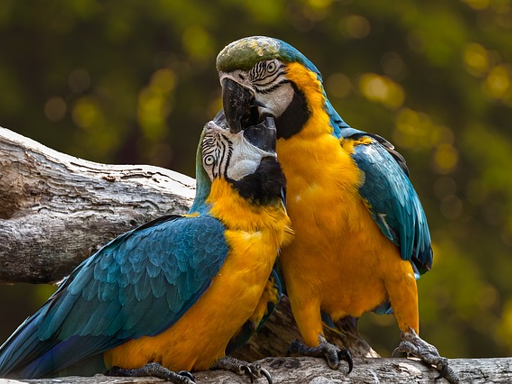}}\hskip.1em
      \subfloat[]{\includegraphics[width=0.15\textwidth]{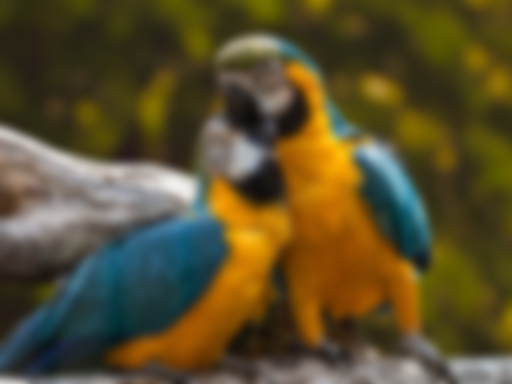}}\hskip.1em
    \subfloat[]{\includegraphics[width=0.15\textwidth]{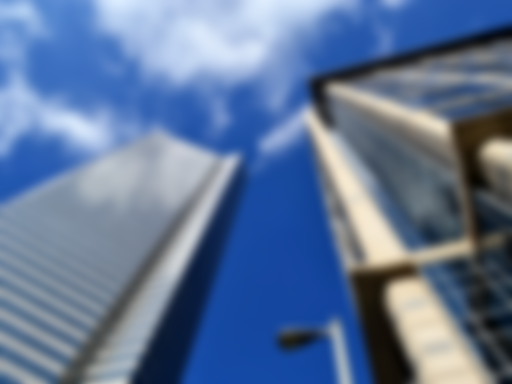}}
    \caption{\textbf{(a)} A ``parrots'' image of pristine quality. \textbf{(b)} A distorted version of (a) by global Gaussian blurring. 
   \textbf{(c)} A distorted ``cityscape'' image by the same level of Gaussian blurring. Humans are able to ``see through'' the Gaussian blur, and recognize the two parrots in (b) with no effort, suggesting the internal representations for the task of visual recognition should be \textit{distortion-insensitive}. This makes it conceptually conflicting to BIQA, which relies on \textit{distortion-sensitive} representations for quality prediction. }\label{fig:example}
\end{figure}

In this paper, we aim to accomplish something in the same spirit, but from a different multitask learning perspective. We ask the key question:

\begin{tightcenter}
    \textit{Can BIQA benefit from auxiliary knowledge provided by other tasks in a multitask learning setting?}
\end{tightcenter}

This question is of particular interest because many high-level computer vision tasks (\eg, object recognition~\cite{deng2009imagenet} and scene classification~\cite{boutell2004learning}), with easier-to-obtain ground-truth labels, seem to be conceptually conflicting to BIQA. This is clearly illustrated in Fig.~\ref{fig:example}. Humans are able to ``see through'' the Gaussian blur, and recognize effortlessly the two parrots in (b). That is, if we would like to develop computational methods for the same purpose, they should rely on \textit{distortion-insensitive} features, and thus be robust to such corruptions. This is also manifested by the common practice in visual recognition that treats synthetic distortions as forms of data augmentation~\cite{hendrycks2019benchmarking}. In stark contrast, BIQA relies preferentially on \textit{distortion-sensitive} features to quantify the perceptual quality of images of various semantic content. Ma \etal~\cite{Ma2018End} proposed a cascaded multitask learning scheme for BIQA, but did not investigate the relationships between BIQA and high-level vision tasks. Fang \etal~\cite{fang2020perceptual} included scene classification as one task, but required manually specifying the parameters (\ie, computations) to share across tasks, which is difficult and bound to be suboptimal.

Taking inspiration from recent work on vision-language pre-training~\cite{radford2021learning}, we propose a general and automated multitask learning scheme for BIQA, with an attempt to answer the above-highlighted question. Here, ``automated'' means that the model parameter sharing for all tasks and the loss weighting assigned to each task are determined automatically. We consider two additional tasks, scene classification and distortion type identification, the former of which is conceptually conflicting to BIQA, while the latter is closely related. We first summarize the scene category, distortion type, and quality level of an input image using a textual template. For example, Fig.~\ref{fig:example} (c) may be described as ``a photo of a \textit{cityscape} with \textit{Gaussian blur} artifacts, which is of \textit{bad} quality." We then employ the contrastive language-image pre-training (CLIP)~\cite{radford2021learning}, a joint vision and language model trained with massive image-text pairs, to obtain the visual and textual embeddings. The joint probability over the three tasks can be computed from the cosine similarities between the image embedding and all candidate textual embeddings\footnote{We specify nine scene categories, eleven distortion types, and five quality levels, giving rise to $495$ textual descriptions/embeddings in total.}. We marginalize the joint distribution to obtain the marginal probability for each task, and further convert the discretized quality levels to a continuous quality score using the marginal distribution as the weighting.

We supplement existing IQA datasets~\cite{sheikh2006statistical, larson2010most, lin2019kadid, ciancio2011no, ghadiyaram2016massive, hosu2020koniq} with scene category and distortion type labels, and jointly optimize the entire method on a combination of them by minimizing a weighted sum of three fidelity losses~\cite{tsai2007frank}, where the loss weightings are adjusted automatically based on the training dynamics~\cite{liu2019end}. From extensive experimental results, we arrive at a positive answer to the highlighted question: BIQA can indeed benefit from both scene classification and distortion type identification. The resulting model, which we name \textbf{L}anguage-\textbf{I}mage \textbf{Q}uality \textbf{E}valuator (LIQE), not only outperforms state-of-the-art BIQA methods~\cite{zhang2020blind, su2020blindly, zhang2021uncertainty, golestaneh2022no} in terms of prediction accuracy on multiple IQA datasets, but also exhibits improved generalizability in the group maximum differentiation (gMAD) competition~\cite{ma2020group}. In addition, we provide quantitative evidence that LIQE better realigns MOSs from different IQA datasets in a common perceptual scale~\cite{perez2020pairwise}.

\section{Related Work}
\label{sec:related}
In this section, we give an overview of recent progress in BIQA, with an emphasis on new paradigms. We then review CLIP~\cite{radford2021learning} and its applications, and  multitask learning in machine learning.
\subsection{BIQA}\label{subsec:related_1}
Conventional BIQA either relied on hand-engineered features in the form of natural scene statistics (NSS)~\cite{moorthy2011blind, mittal2012no, mittal2013making, ghadiyaram2017perceptual} or shallow feature  learning\cite{ye2012unsupervised,kang2014convolutional,xu2016blind} in the form of codebooks. Deep learning takes advantage of the end-to-end optimization of feature extraction and quality regression, and has significantly advanced the field of BIQA. Over the years, BIQA methods have explored different \textit{computational structures}: generalized divisive normalization~\cite{Ma2018End}, multi-level feature aggregation~\cite{li2019has},  adaptive convolution~\cite{su2020blindly}, and self-attention~\cite{ke2021musiq,golestaneh2022no, yang2022maniqa}; and \textit{objective functions}: $\ell_p$-norm induced metric, fidelity-based ranking losses~\cite{tsai2007frank}, Pearson linear correlation coefficient (PLCC), and differentiable approximations to Spearman rank correlation coefficient (SRCC)~\cite{blondel2020fast}, and generalized correlation loss for accelerated convergence~\cite{li2020norm}.

 \begin{figure*}[t]
    \centering
    \includegraphics[width=.8\textwidth]{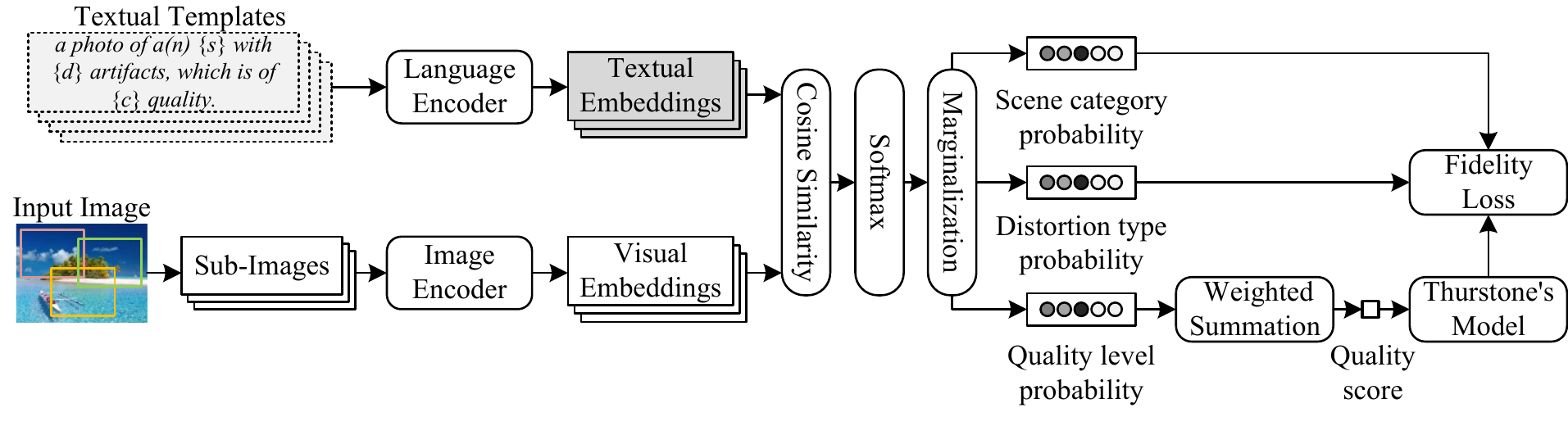}
    \caption{System diagram of the proposed LIQE.}\label{fig:overview}
\end{figure*}

New paradigms for BIQA flourish in  recent years, aiming at exploring promising directions of next-generation BIQA. Representative work includes: patch-to-picture learning for local quality prediction~\cite{ying2020from}, active learning for worthy sample identification~\cite{wang2022active, Wang2021troubleshooting}, unified optimization for cross-distortion scenarios~\cite{zhang2021uncertainty}, meta-learning for fast adaptation~\cite{zhu2020metaiqa}, continual learning for streaming distortions ~\cite{zhang2023continual, liu2022liqa, ma2021remember, zhang2021task}, and perceptual attacks for robustness evaluation~\cite{zhang2022perceptual}. In this paper, we leverage multitask learning to facilitate auxiliary knowledge transfer.

\subsection{CLIP Applications}\label{subsec:related_2}
 CLIP has shown great promise to assist a broad scope of vision tasks. Originally, Radford~\etal~\cite{radford2021learning} leveraged $400$ million image-text pairs to pre-train a family of CLIP models, which present remarkable zero-shot transfer ability to a wide range of downstream vision tasks. Zhou~\etal~\cite{zhou2022learning} suggested prompt tuning to improve transfer effectiveness, at the cost of language interpretability. Shortly after its inception, CLIP has found its way to (open-vocabulary) semantic segmentation~\cite{li2022language,xu2022groupvit} and object detection~\cite{gu2022open,li2022grounded}. Vinker~\etal~\cite{vinker2022clipasso} found a novel use of  CLIP for object sketching with excellent understanding of object semantics. 

Closest to ours, Wang~\etal~\cite{wang2022exploring} assumed CLIP models to be inherently quality-aware, and adopted them to assess image quality and aesthetics via prompt engineering. Our method differs significantly from theirs~\cite{wang2022exploring} both conceptually and computationally. We exploit CLIP in the multitask learning setting to aid BIQA by  auxiliary knowledge transfer. Moreover, we fine-tune the CLIP model instead of prompt tuning for much better quality prediction performance without sacrificing language interpretability.

\subsection{Multitask Learning}\label{subsec:related_3}
Multitask learning as a specific case of multi-objective optimization~\cite{marler2004survey} typically improves task accuracy, memory cost, and inference time through shared computation and information across tasks. Existing methods differ mainly in two design choices: model parameter sharing and loss weighting. Instead of manually specifying which parameters to share~\cite{argyriou2006multi, misra2016cross,kokkinos2017ubernet} or learning to determine task specific parameters with a combinatorial complexity~\cite{ruder2019latent, VandenhendeGGB20, MallyaDL18, sun2020adashare, wallingford2022task}, we assume all parameters in the image encoder of LIQE are shared, whose capacity is dynamically allocated to each task during end-to-end optimization. For loss weighting,  Sener~\etal~\cite{sener2018multi} and Lin~\etal~\cite{lin2019pareto} cast multitask learning as multi-objective optimization, and solved for the optimal loss weighting according to the Karush-Kuhn-Tucher conditions. Liu~\etal~\cite{liu2022auto} implemented an automated weighting scheme based on model agnostic meta learning~\cite{finn2017model}, which allows specification of primary (\ie, the most important) tasks. Other effective heuristics for loss weighting include learning task uncertainty~\cite{KendallGC18}, gradient normalization~\cite{chen2018gradnorm}, and loss descending rate~\cite{liu2019end}. In this paper, we adopt the method in~\cite{liu2019end} due to its conceptual simplicity, computational convenience, and high efficiency.

\section{LIQE from Multitask Learning}
\label{sec:proposed}
In this section, we first present some preliminaries of the problem formulation. We then describe a general and automated multitask learning scheme for BIQA based on vision-language correspondence, followed by specifications of loss functions to drive the end-to-end optimization. The system diagram of LIQE is shown in Fig.~\ref{fig:overview}. 
\subsection{Preliminaries}
Given an image $\bm{x}\in\mathbb{R}^N$  that may undergo several stages of degradations, the goal of a BIQA model $\hat{q}:\mathbb{R}^N\mapsto\mathbb{R}$ is to predict the perceptual quality of $\bm{x}$, close to its MOS $q(\bm{x})\in\mathbb{R}$. We also work with a Likert-scale of five quality levels~\cite{sheikh2006statistical, ghadiyaram2016massive}: $c \in\mathcal{C}=\{1,2,3,4,5\} =$ \{``bad", ``poor", ``fair”, ``good”, ``perfect"\}, and relate $c$ to $\hat{q}$ by
\begin{align}\label{eq:stl}
    \hat{q}(\bm{x}) = \sum_{c=1}^{C}\hat{p}(c|\bm{x})\times c,
\end{align}
where $C = 5$ is the number of quality levels and $\hat{p}(c|\cdot)$ is the marginal probability of $c$ to be estimated. 

Apart from BIQA, we also include a conceptually conflicting task - scene classification and a closely related task - distortion type identification.   We consider nine scene categories: $s\in\mathcal{S}=$ \{``animal", ``cityscape", ``human", ``indoor scene", ``landscape", ``night scene", ``plant", ``still-life", and ``others"\}. An image may contain multiple scene labels. We do not discriminate between synthetic and realistic distortions, and instead identify the dominant distortion in the image: $d\in\mathcal{D}=$ \{``blur", ``color-related", ``contrast", ``JPEG compression", ``JPEG2000 compression'', ``noise", ``over-exposure", ``quantization", ``under-exposure", ``spatially-localized", and ``others"\} with eleven in total. According to our characterization, the ``others'' category includes images with no distortions (\ie, of pristine quality). It is then natural to create a textual template to put together labels from the three tasks: ``\textit{a photo of a(n) $\{s\}$ with $\{d\}$ artifacts, which is of $\{c\}$ quality}'', and we have $5\times 9\times 11=495$ candidate textual descriptions.

\subsection{Vision-Language Correspondence}
\noindent\textbf{Joint Probability over Multiple Tasks}. The proposed LIQE relies on a pre-trained CLIP model~\cite{radford2021learning} for feature embeddings. CLIP consists of an image encoder $\bm{f}_{\bm{\phi}}:\mathbb{R}^N\mapsto\mathbb{R}^K$ and a language encoder $\bm{g}_{\bm{\varphi}}: \mathcal{T} \mapsto \mathbb{R}^K$, parameterized by $\bm{\phi}$ and $\bm{\varphi}$, respectively, where $\mathcal{T}$ denotes the text corpus.  Note that $\bm{f}_{\bm{\phi}}$ and $\bm{g}_{\bm{\varphi}}$ share the same feature dimension by design. As also part of the model design, $\bm{f}_{\bm{\phi}}$ accepts input images with a fixed spatial size (\ie, $224 \times 224 \times 3$). A na\"{i}ve resizing (\eg, by bilinear interpolation) may change the perceptual quality, which corresponds to adjusting the effective viewing distance. As a result, we prefer cropping $U$ sub-images from $\bm{x}$, and obtain the visual embedding matrix  $\bm{F}({\bm{x}})\in \mathbb{R}^{U \times K}$. Similarly, given a total of  $V$ candidate textual descriptions by the lower-cased byte pair encoding (BPE)~\cite{SennrichHB16a}, we use the language encoder to obtain the textual embedding matrix $\bm{G}(\bm{x}) \in \mathbb{R}^{V \times K}$, where $V=495$. 

We then compute the cosine similarity between the visual embedding of the $u$-th sub-image $\bm{F}_{u\bullet}$ (as a row vector and for $1\le u \le U$) and the $v$-th candidate textual embedding $\bm{G}_{v\bullet}$ (corresponding to a particular set of $\{c, s, d\}$), averaging across $U$ sub-images to obtain the image-level correspondence score:
\begin{align}\label{eq:logit}
    \mathrm{logit}(c, s, d|\bm{x})=\frac{1}{U}\sum_{u=1}^{U}\frac{\bm{F}_{u\bullet}(\bm{x})\bm{G}^\intercal_{v\bullet}(\bm{x})}{\|\bm{F}_{u\bullet}(\bm{x})\|_2\|\bm{G}_{v\bullet}(\bm{x})\|_2}.
\end{align}

After corresponding the image to all candidate textual descriptions, we apply a softmax function to compute a joint probability with a learnable temperature parameter $\tau_1$:
\begin{align}\label{eq:jointp}
    \hat{p}(c, s, d|\bm{x}) = \frac{\exp\left(\mathrm{logit}(c,s,d|\bm{x})/\tau_1\right)}{\sum_{c,s,d}\exp\left(\mathrm{logit}(c,s,d|\bm{x})/\tau_1\right)}.
\end{align}
Ideally, we shall minimize the statistical distance between the predicted joint probability and the ground-truth joint probability $p(c, s, d|\bm{x})$ for optimizing the parameter vector $\bm{\theta}=\{\bm{\phi},\bm{\varphi},\tau\}$. However, $p(c, s, d|\bm{x})$ may not be inferred accurately due to the fact that  existing IQA datasets only provide continuous quality scores instead of discrete quality levels. Moreover, different IQA datasets have different perceptual scales due to differences in subjective testing~\cite{zhang2021uncertainty}, which further complicates quality score-to-level conversion.

\noindent\textbf{Loss for BIQA}. 
Given $\hat{p}(c,s,d|\bm{x})$, we marginalize it to obtain $\hat{p}(c|\bm{x})$, and compute the quality estimate $\hat{q}(\bm{x})\in\mathbb{R}$ by Eq.~\eqref{eq:stl}. We  consider the pairwise learning-to-rank model estimation for BIQA. Specifically, for an image pair $(\bm{x}, \bm{y})$ from the same IQA dataset,  we compute a binary label according to their ground-truth MOSs:
\begin{align}\label{eq:bgt}
     p(\bm{x},\bm{y}) = 
\begin{cases} 
 1 & \mbox{if } q(\bm{x})\ge q(\bm{y}) \\
      0 & \mbox{otherwise} \end{cases}.
\end{align}
Under the Thurstone's model~\cite{thurstone1927law}, we estimate the probability of $\bm{x}$  perceived better than $\bm{y}$ as
\begin{align}\label{eq:thurstone}
\hat{p}(\bm{x}, \bm{y})= \Phi\left(\frac{\hat{q}(\bm{x}) - \hat{q}(\bm{y})}{\sqrt{2}}\right),
\end{align}
where $\Phi(\cdot)$ is the standard Normal cumulative distribution function, and the variance is fixed to one. We adopt the fidelity loss~\cite{tsai2007frank} as the statistical distance measure:
\begin{align}\label{eq:fidelity}
\ell_{q}(\bm{x},\bm{y};\bm{\theta})&
= 1 - \sqrt{p(\bm{x}, \bm{y})\hat{p}(\bm{x}, \bm{y})} \nonumber \\ &-\sqrt{(1-p(\bm{x}, \bm{y}))(1-\hat{p}(\bm{x}, \bm{y}))}.
\end{align}

\noindent \textbf{Loss for Scene Classification}. In our setting, an image can be assigned to one or more scene categories, leading to a multi-label classification problem. Thus, we compute an average of $S$ binary fidelity losses:
\begin{align}\label{eq:fidelity_scene}
\ell_{s}(\bm{x};\bm{\theta})
=&\frac{1}{\vert\mathcal{S}\vert}\sum_{s\in \mathcal{S}}\biggl(1 - \sqrt{p(s|\bm{x})\hat{p}(s|\bm{x})} \nonumber \\ &-\sqrt{(1-p(s|\bm{x}))(1 - \hat{p}(s|\bm{x}))}\biggr),
\end{align}
where $p(s|\bm{x}) = 1$ if $\bm{x}$ is in the $s$ category and zero otherwise. $\sum_sp(s|\bm{x}) =S$, where $1\le S \le \vert\mathcal{S}\vert$ is the number of target categories to which $\bm{x}$ belongs. $\hat{p}(s|\bm{x})$ is the marginal probability computed from Eq.~\eqref{eq:jointp}. We also try a softmax loss formulation:
\begin{align}\label{eq:fidelity_scene2}
\ell_{s}(\bm{x};\bm{\theta})
=&\left(1 - \sum_{s\in \mathcal{S}}\sqrt{p(s|\bm{x})\hat{p}(s|\bm{x})}\right),
\end{align}
where equal probabilities are assigned to the target scene categories with $\sum_s p(s|\bm{x})=1$. Similar results are obtained, which we attribute to the relatively simple setting of our scene classification problem with only nine categories. 

\begin{table*}[t]
  \centering
  \caption{Median SRCC and PLCC results across ten sessions along with the standard deviation in the bracket. CLIVE stands for the LIVE Challenge Database. The top two results are highlighted in bold.}\label{tab:overall}
  \begin{adjustbox}{max width=\textwidth}
  \begin{tabular}{l|cccccc}
      \toprule
    Dataset & {LIVE~\cite{sheikh2006statistical}} & {CSIQ~\cite{larson2010most}} & {KADID-10k~\cite{lin2019kadid}} & {BID~\cite{ciancio2011no}} &{CLIVE~\cite{ghadiyaram2016massive}} & {KonIQ-10k~\cite{hosu2020koniq}}\\
    \hline
    Criterion & \multicolumn{6}{c}{SRCC} \\
     \hline
        NIQE & 0.908 ({\color{grey}$\pm$ 0.017}) & 0.631 ({\color{grey}$\pm$ 0.038}) & 0.389 ({\color{grey}$\pm$ 0.019}) & 0.573 ({\color{grey}$\pm$ 0.044}) & 0.446 ({\color{grey}$\pm$ 0.065}) & 0.415 ({\color{grey}$\pm$ 0.019}) \\
       ILNIQE & 0.887 ({\color{grey}$\pm$ 0.032})  & 0.808 ({\color{grey}$\pm$ 0.039}) & 0.565 ({\color{grey}$\pm$ 0.013})  & 0.548 ({\color{grey}$\pm$ 0.044})  & 0.469 ({\color{grey}$\pm$ 0.063})  & 0.509 ({\color{grey}$\pm$ 0.021}) \\
        Ma19 & 0.922 ({\color{grey}$\pm$ 0.024}) & 0.926 ({\color{grey}$\pm$ 0.017}) & 0.465 ({\color{grey}$\pm$ 0.019})  & 0.373 ({\color{grey}$\pm$ 0.059})  & 0.336 ({\color{grey}$\pm$ 0.038})  & 0.360 ({\color{grey}$\pm$ 0.013})\\

      PaQ2PiQ & 0.544 ({\color{grey}$\pm$ 0.033}) & 0.697 ({\color{grey}$\pm$ 0.040}) & 0.403 ({\color{grey}$\pm$ 0.021}) & 0.719 ({\color{grey}$\pm$ 0.043}) & 0.732 ({\color{grey}$\pm$ 0.036}) & 0.722 ({\color{grey}$\pm$ 0.012})\\
      KonCept & 0.673 ({\color{grey}$\pm$ 0.040}) & 0.631 ({\color{grey}$\pm$ 0.064}) & 0.503 ({\color{grey}$\pm$ 0.025}) & 0.816 ({\color{grey}$\pm$ 0.029}) & 0.778 ({\color{grey}$\pm$ 0.024}) & 0.911 ({\color{grey}$\pm$ 0.005})\\
      MUSIQ & 0.837 ({\color{grey}$\pm$ 0.011})& 0.697 ({\color{grey}$\pm$ 0.040}) & 0.572 ({\color{grey}$\pm$ 0.027}) & 0.744  ({\color{grey}$\pm$ 0.038})& 0.785 ({\color{grey}$\pm$ 0.029})& {\bf 0.915} ({\color{grey}$\pm$ 0.003})\\

      DBCNN & 0.963 ({\color{grey}$\pm$ 0.012})& {\bf 0.940} ({\color{grey}$\pm$ 0.015})& 0.878 ({\color{grey}$\pm$ 0.023}) & {\bf 0.864}  ({\color{grey}$\pm$ 0.016})& 0.835 ({\color{grey}$\pm$ 0.022})& 0.864 ({\color{grey}$\pm$ 0.007})\\
      HyperIQA & {\bf 0.966} ({\color{grey}$\pm$ 0.012}) & 0.934 ({\color{grey}$\pm$ 0.031}) & 0.872 ({\color{grey}$\pm$ 0.017}) & 0.848 ({\color{grey}$\pm$ 0.033}) & {\bf 0.855} ({\color{grey}$\pm$ 0.021}) & 0.900 ({\color{grey}$\pm$ 0.006})\\
      TreS & 0.965 ({\color{grey}$\pm$ 0.019})& 0.902 ({\color{grey}$\pm$ 0.041}) & 0.881({\color{grey}$\pm$ 0.019}) & 0.853  ({\color{grey}$\pm$ 0.031})& 0.846 ({\color{grey}$\pm$ 0.020})& 0.907 ({\color{grey}$\pm$ 0.007})\\
       UNIQUE& 0.961 ({\color{grey}$\pm$ 0.005}) & 0.902 ({\color{grey}$\pm$ 0.052}) & {\bf 0.884} ({\color{grey}$\pm$ 0.013}) & 0.852 ({\color{grey}$\pm$ 0.027}) & 0.854 ({\color{grey}$\pm$ 0.020}) & 0.895 ({\color{grey}$\pm$ 0.008})\\
       \hline
       LIQE & {\bf 0.970} ({\color{grey}$\pm$ 0.004}) & {\bf 0.936} ({\color{grey}$\pm$ 0.025}) & {\bf 0.930} ({\color{grey}$\pm$ 0.009}) & {\bf 0.875} ({\color{grey}$\pm$ 0.020}) & {\bf 0.904} ({\color{grey}$\pm$ 0.014}) & {\bf 0.919} ({\color{grey}$\pm$ 0.004})\\
    \midrule
    Criterion  & \multicolumn{6}{c}{PLCC}\\
     \hline
        NIQE & 0.904 ({\color{grey}$\pm$ 0.089}) & 0.719 ({\color{grey}$\pm$ 0.022}) & 0.442 ({\color{grey}$\pm$ 0.019}) & 0.618 ({\color{grey}$\pm$ 0.045}) & 0.507 ({\color{grey}$\pm$ 0.054}) & 0.438 ({\color{grey}$\pm$ 0.015})\\
       ILNIQE & 0.894 ({\color{grey}$\pm$ 0.025}) & 0.851 ({\color{grey}$\pm$ 0.029}) & 0.611 ({\color{grey}$\pm$ 0.027})& 0.494 ({\color{grey}$\pm$ 0.046}) & 0.518 ({\color{grey}$\pm$ 0.051}) &0.534 ({\color{grey}$\pm$ 0.020})\\
        Ma19 & 0.923 ({\color{grey}$\pm$ 0.019}) & 0.929 ({\color{grey}$\pm$ 0.011}) & 0.501 ({\color{grey}$\pm$ 0.013}) & 0.399 ({\color{grey}$\pm$ 0.059}) & 0.405 ({\color{grey}$\pm$ 0.033})& 0.398 ({\color{grey}$\pm$ 0.009}) \\
 
      PaQ2PiQ & 0.558 ({\color{grey}$\pm$ 0.023}) & 0.766 ({\color{grey}$\pm$ 0.028}) & 0.448 ({\color{grey}$\pm$ 0.012}) & 0.700 ({\color{grey}$\pm$ 0.032}) & 0.755 ({\color{grey}$\pm$ 0.022}) & 0.716 ({\color{grey}$\pm$ 0.009})\\
      KonCept& 0.619 ({\color{grey}$\pm$ 0.039}) & 0.645 ({\color{grey}$\pm$ 0.043}) & 0.515 ({\color{grey}$\pm$ 0.018}) & 0.825 ({\color{grey}$\pm$ 0.026}) & 0.799 ({\color{grey}$\pm$ 0.016}) & {\bf 0.924} ({\color{grey}$\pm$ 0.003})\\
      MUSIQ & 0.818 ({\color{grey}$\pm$ 0.011})& 0.766 ({\color{grey}$\pm$ 0.028}) & 0.584 ({\color{grey}$\pm$ 0.016}) & 0.774  ({\color{grey}$\pm$ 0.030})& 0.828 ({\color{grey}$\pm$ 0.017})& {\bf 0.937} ({\color{grey}$\pm$ 0.003})\\

      DBCNN & \textbf{0.966} ({\color{grey}$\pm$ 0.010})& {\bf 0.954} ({\color{grey}$\pm$ 0.013})& 0.878 ({\color{grey}$\pm$ 0.022}) & {\bf 0.883}  ({\color{grey}$\pm$ 0.017})& 0.854 ({\color{grey}$\pm$ 0.015})& 0.868 ({\color{grey}$\pm$ 0.006})\\
      HyperIQA & {\bf 0.968} ({\color{grey}$\pm$ 0.011})& {\bf 0.946} ({\color{grey}$\pm$ 0.022}) & 0.869 ({\color{grey}$\pm$ 0.018}) & 0.868  ({\color{grey}$\pm$ 0.027})& 0.878 ({\color{grey}$\pm$ 0.015})& 0.915 ({\color{grey}$\pm$ 0.004})\\
      TreS & 0.963 ({\color{grey}$\pm$ 0.016})& 0.923  ({\color{grey}$\pm$ 0.031})& 0.879 ({\color{grey}$\pm$ 0.019}) & 0.871  ({\color{grey}$\pm$ 0.028})& 0.877 ({\color{grey}$\pm$ 0.016})& {\bf 0.924} ({\color{grey}$\pm$ 0.006})\\

       UNIQUE& 0.952 ({\color{grey}$\pm$ 0.007})  & 0.921 ({\color{grey}$\pm$ 0.048})&  {\bf 0.885} ({\color{grey}$\pm$ 0.011}) & 0.875 ({\color{grey}$\pm$ 0.019}) & {\bf 0.884} ({\color{grey}$\pm$ 0.014}) & 0.900 ({\color{grey}$\pm$ 0.005})\\
       \hline
       LIQE & 0.951 ({\color{grey}$\pm$ 0.006})  &  0.939 ({\color{grey}$\pm$ 0.024})& {\bf 0.931} ({\color{grey}$\pm$ 0.009}) & {\bf 0.900} ({\color{grey}$\pm$ 0.016}) & {\bf 0.910} ({\color{grey}$\pm$ 0.013}) & 0.908 ({\color{grey}$\pm$ 0.002})\\
     \bottomrule
   \end{tabular}
   \end{adjustbox}
\end{table*}

\noindent \textbf{Loss for Distortion Type Identification}. Because we only consider the dominant distortion type in  $\bm{x}$, distortion type identification can be formulated as a standard multi-class classification problem. We use the multi-class fidelity loss:
\begin{align}\label{eq:fidelity_dis}
\ell_{d}(\bm{x};\bm{\theta})
=&\left(1 - \sum_{d\in \mathcal{D}}\sqrt{p(d|\bm{x})\hat{p}(d|\bm{x})}\right),
\end{align}
where $p(d|\bm{x}) = 1$ if $\bm{x}$ contains the $d$ artifacts, and zero otherwise. $\hat{p}(d|\bm{x})$ is computed similarly by marginalizing the joint probability in Eq.~\eqref{eq:jointp}.

\noindent\textbf{Final Loss for Multitask Learning}. Similar to~\cite{zhang2021uncertainty}, we choose to jointly train a BIQA model on $M$ ($M \geq 2$) IQA datasets. At the $t$-th training iteration and from the $m$-th dataset, we sample a mini-batch $\mathcal{B}^{(m)}_t$, and form all possible pairs of images with ground-truths (by Eq.~\eqref{eq:bgt}), which are collectively denoted by $\mathcal{P}^{(m)}_t$. We combine $\{\mathcal{B}^{(m)}_t\}_{i=1}^M$ and $\{\mathcal{P}^{(m)}_t\}_{i=1}^M$ to form $\mathcal{B}_t$ and $\mathcal{P}_t$, respectively, based on which we define the final loss as a linearly weighted summation of the three individual losses:
\begin{align}\label{eq:fl}
 \ell(\mathcal{B},t;\bm{\theta}) 
&=\frac{1}{\vert\mathcal{P}\vert}\sum_{(\bm{x}, \bm{y})\in\mathcal{P}}\lambda_q(t)\ell_q(\bm{x},\bm{y};\bm{\theta})+ \nonumber \\ 
&\frac{1}{\vert\mathcal{B}\vert}\sum_{\bm{x}\in\mathcal{B}}\big(\lambda_s(t)\ell_s(\bm{x}) + \lambda_d(t)\ell_d(\bm{x})\big).
\end{align}
The weighting vector $\bm{\lambda}(t) = [\lambda_q(t),\lambda_s(t),\lambda_d(t)]^\intercal$ at the $t$-th iteration can be automatically computed according to the relative descending rate~\cite{liu2019end}:
\begin{align}\label{eq:dwa}
\lambda_{j}(t) =\frac{\exp\left(w_j(t-1)/\tau_2\right)}{\sum_i\exp\left(w_i(t-1)/\tau_2\right)}, w_j(t-1) = \frac{\ell_{j}(t - 1)}{\ell_{j}(t - 2)},
\end{align}
where $i,j \in \{q,s,d\}$, and $\tau_2$ is another fixed temperature parameter. When $\tau_2$ is sufficiently large, we approach a uniform distribution, where different losses are weighted equally. In our experiments, we follow the suggestions in~\cite{liu2019end}, and compute $w_j(t-1)$ through the moving average loss over the recent iterations in each epoch.

\section{Experiments}
\label{sec:exp}
In this section, we first present the experimental setups, upon which we compare LIQE with several state-of-the-art BIQA methods. We then evaluate LIQE's ability to realign MOSs from different datasets qualitatively and quantitatively. The rationality of each design choice is verified through a series of ablation studies. Finally, we provide an analysis of task relationships.
\subsection{Experimental Setups}\label{subsec:exp_setup}
We conduct experiments on six IQA datasets, among which LIVE~\cite{sheikh2006statistical}, CSIQ~\cite{larson2010most}, and KADID-10k~\cite{lin2019kadid} contain synthetic distortions, while LIVE Challenge~\cite{ghadiyaram2016massive} (denoted as CLIVE in Table~\ref{tab:overall}), BID~\cite{ciancio2011no}, and KonIQ-10K~\cite{hosu2020koniq} include realistic distortions. We randomly sample $70\%$ and $10\%$ images from each dataset to construct the training and validation set, respectively, leaving the remaining $20\%$ for testing. Regarding the three datasets with synthetic distortions, we split the train/val/test sets according to the reference images in order to ensure content independence. We repeat this procedure ten times, and report median SRCC and PLCC results as prediction monotonicity and precision measures, respectively.

We adopt ViT-B/32~\cite{radford2021learning} as the visual encoder and GPT-2~\cite{radford2019language} with a base size of 63M parameters as the text encoder. We train the model by minimizing the objective in Eq.~\eqref{eq:fl} using AdamW~\cite{LoshchilovH19} with a decoupled weight decay regularization of $10^{-3}$. The initial learning rate is set to $5\times10^{-6}$, which is scheduled by a cosine annealing rule~\cite{LoshchilovH17}. We optimize LIQE for $80$ epochs with a mini-batch size of $4$ on the LIVE, CSIQ, BID, and LIVE Challenge datasets, and size of $16$ on KonIQ-10k and KADID-10k. During training and inference, we randomly crop $3$ and $15$ sub-images with a spatial size of $224 \times 224 \times 3$ from original images without changing their aspect ratios. All experiments are conducted on a single NVIDIA GeForce RTX 3090 GPU.

\subsection{Main Results}\label{subsec:main_result}
We compare the performance of the proposed LIQE against three opinion-unaware BIQA models, including 
NIQE~\cite{mittal2013making}, ILNIQE~\cite{zhang2015feature}, and Ma19~\cite{ma2019blind}, and seven data-driven DNN-based methods, including 
PaQ2PiQ~\cite{ying2020from}, KonCept~\cite{hosu2020koniq}, MUSIQ~\cite{ke2021musiq}, DBCNN~\cite{zhang2020blind}, HyperIQA~\cite{su2020blindly}, TreS~\cite{golestaneh2022no}, and UNIQUE~\cite{zhang2021uncertainty}. For competing models, we either directly adopt the publicly available implementations~\cite{bosse2016deep, hosu2020koniq, ying2020from, ke2021musiq}, or re-train them on our datasets with 
the training codes provided by the respective authors~\cite{zhang2020blind, su2020blindly, golestaneh2022no, zhang2021uncertainty}. DBCNN, HyperIQA, and TreS are separately trained on each individual dataset, while UNIQUE is jointly trained on six datasets. We summarize the median SRCC and~PLCC results across ten sessions in Table~\ref{tab:overall}, from which we draw several insightful observations. First, although opinion-unaware methods aim ambitiously for handling arbitrary distortions, they only perform well on two legacy datasets LIVE~\cite{sheikh2006statistical} and CSIQ~\cite{larson2010most}, which contain 
the most basic distortions in the field of IQA. Second, despite being trained on a relatively large-scale dataset, PaQ2PiQ~\cite{ying2020from} only presents reasonable quality prediction accuracy on datasets with realistic distortions~\cite{ciancio2011no, ghadiyaram2016massive, hosu2020koniq}. Equipped with more sophisticated backbone networks and trained on KonIQ-10k~\cite{hosu2020koniq}, KonCept~\cite{hosu2020koniq} and MUSIQ~\cite{ke2021musiq} deliver better performance. However, neither of them presents promising 
generalization capabilities
towards complex synthetic distortions in KADID-10k~\cite{lin2019kadid}, demonstrating the challenges in
handling cross-distortion scenarios.

Being separately trained on each individual dataset with a separate group of parameters, DBCNN~\cite{zhang2020blind}, HyperIQA~\cite{su2020blindly}, and TreS~\cite{golestaneh2022no} achieve promising performance on all six datasets. The pairwise learning-to-rank training strategy enables UNIQUE~\cite{zhang2021uncertainty} and LIQE to learn from multiple datasets simultaneously with a single set of weights. Their competitive performance against the separately trained BIQA methods is clearly demonstrated.
Moreover, LIQE outperforms UNIQUE with clear margins on the three datasets with realistic distortions~\cite{ciancio2011no, ghadiyaram2016massive, hosu2020koniq} and KADID-10k~\cite{lin2019kadid} (commonly regarded as the most challenging BIQA benchmark with synthetic distortions), which verifies the effectiveness of the proposed multitask learning scheme based on the vision-language correspondence.

\subsection{Cross-Dataset Evaluation}\label{subsec:crossdataset}
We compare the generalizability of LIQE against competitive BIQA models in a more challenging cross-dataset setting. Specifically, we employ the full TID2013~\cite{ponomarenko2013color} and SPAQ~\cite{fang2020perceptual} as the test sets, which contain synthetic and realistic camera distortions, respectively. In addition, we also test BIQA models using images from the training set\footnote{We are intended to participate in the NTIRE challenge competition, where evaluations on the validation and test sets can only be performed online by registered participants.} of PIPAL, whose MOSs are publicly available~\cite{jinjin2020pipal}. PIPAL gathers and annotates images enhanced  by various image restoration algorithms. The introduced algorithm-dependent distortions, especially those arising from generative adversarial network (GAN)-based methods, appear quite different from those covered in the six datasets described in Sec.~\ref{subsec:exp_setup}. It is clear from Table~\ref{tab:cross} that only UNIQUE~\cite{zhang2021uncertainty} and LIQE are capable of handling both synthetic and realistic distortions well with single sets of parameters, which highlights the promise of the joint training on multiple datasets.
Due to the significant distributional shifts in distortion patterns, none of the tested methods presents promising results on PIPAL~\cite{jinjin2020pipal}, suggesting that task-specific model training may be a viable option to handle such algorithm-dependent distortions. Nevertheless, LIQE achieves a slightly higher SRCC result, which we believe is attributed to the better commonsense knowledge of visual quality learned from the vision-language correspondence.

\begin{figure*}[t]
    \centering
    \includegraphics[width=\textwidth]{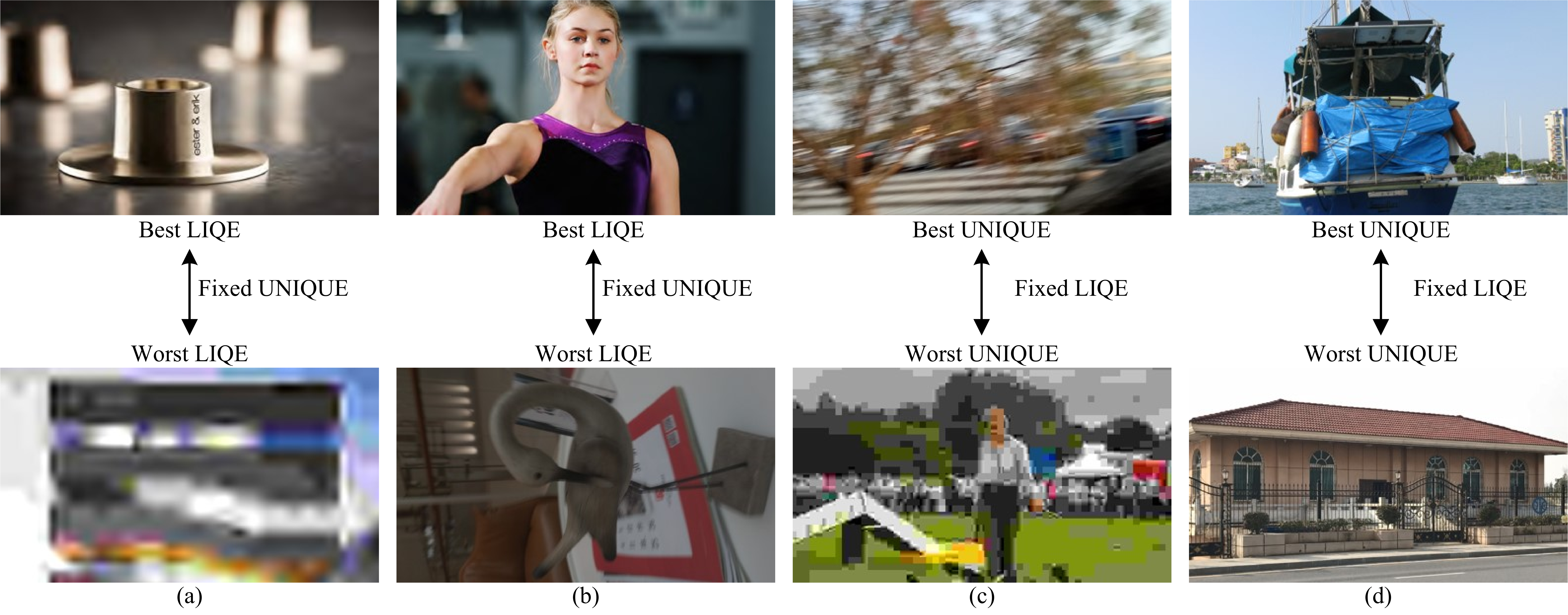}
    \caption{gMAD competition results between UNIQUE~\cite{zhang2021uncertainty} and LIQE. \textbf{(a)} Fixed UNIQUE at the low-quality level. \textbf{(b)} Fixed UNIQUE at the high-quality level. \textbf{(c)} Fixed LIQE at the low-quality level. \textbf{(d)} Fixed LIQE at the high-quality level. }\label{fig:gMAD}
\end{figure*}

\begin{table}[tp]
  \centering
  \caption{SRCC results on the three IQA datasets under the cross-dataset setup. The subscripts ``$s$'' and ``$r$'' stand for models trained on KADID-10K~\cite{lin2019kadid} and KonIQ-10K~\cite{hosu2020koniq}, respectively. Best results are highlighted in bold.}\label{tab:cross}

  \begin{tabular}{l|ccc}
      \toprule
    Dataset & {TID2013~\cite{ponomarenko2013color}} & {SPAQ~\cite{fang2020perceptual}} & {PIPAL ~\cite{jinjin2020pipal}}\\
    \hline
     NIQE & 0.314 & 0.578 & 0.153\\
    DBCNN$_{r}$ & 0.471 & 0.801 & 0.413 \\
    DBCNN$_{s}$ & 0.686 & 0.412 & 0.321 \\
    PaQ2PiQ & 0.423 & 0.823 & 0.400 \\
    MUSIQ$_{r}$ & 0.584 & 0.853 & 0.450 \\
    \hline
    UNIQUE & 0.768 & 0.838 & 0.444\\
    LIQE & \textbf{0.811} & \textbf{0.881} & \textbf{0.478}\\
     \bottomrule
   \end{tabular}
\end{table}

\begin{table}[t]
  \centering
  \caption{Median SRCC results across ten sessions in the multi-dataset realignment experiment.}\label{tab:align}

  \begin{tabular}{l|cc}
      \toprule
    Method & UNIQUE & LIQE\\
    \hline
    SRCC & 0.851 & 0.879\\
     \bottomrule
   \end{tabular}
\end{table}

\begin{table*}[t]
  \centering
  \caption{Median SRCC results of LIQE variants across ten sessions.  The top results are highlighted in bold.}\label{tab:ablation}
\begin{adjustbox}{max width=\textwidth}
  \begin{tabular}{l|ccccccc}
      \toprule
    Variants & {LIVE} & {CSIQ} & {KADID-10k} & {BID} &{CLIVE} & {KonIQ-10k} & Mean\\
     \hline
    (1) Pretrained CLIP ($C=2$) & 0.801 &  0.777 & 0.642 & 0.668 & 0.595 & 0.677 &0.693\\   
    (2) Fine-Tuned CLIP ($C=2$) & 0.934 &  0.900 & 0.901 & 0.841 & 0.868 & 0.898 & 0.890 \\    
    (3) Frozen Textual Encoder $\bm{g}_{\bm{\varphi}}$ & 0.968 &  0.933 & 0.926 & 0.866 & 0.901 & 0.916 & 0.918 \\    
    (4) Separate Task Templates& 0.969 &  0.937 & 0.924 & 0.856 & 0.903 & 0.905 & 0.916\\    
    (5) Equal Task Weightings & 0.968 &  \textbf{0.938} & 0.927 & 0.870 & 0.899 & 0.917 & 0.920 \\ 

    (6) Fine-Tuned Visual Encoder $\bm{f}_{\bm{\phi}}$ Only & 0.969 &  0.923 & 0.921 & 0.863 & 0.901 & 0.910 & 0.915 \\  
    \hline
    LIQE& {\bf 0.970} &  0.936 & {\bf 0.930} & {\bf 0.875} & {\bf 0.904} & {\bf 0.919} & \textbf{0.922} \\
     \bottomrule
   \end{tabular}
   \end{adjustbox}
\end{table*}

\begin{table}[htp]
  \centering
  \caption{Median mean (mSRCC) and  mean accuracy (mACC) results of LIQE variants across ten sessions, trained on different task combinations of the six datasets. The subscripts ``$s$'' and ``$r$'' stand for the accuracy of scene classification and distortion type identification, respectively.}\label{tab:relation}
  \begin{tabular}{l|ccc}
      \toprule
    Task Combination & SRCC & ACC$_{s}$ & ACC$_d$\\
    \hline
    Quality & 0.915 & -- & --\\
    Scene & -- & 0.908 & --\\
    Distortion & -- & -- & 0.847\\
    Quality + Scene & 0.915 & 0.898 & --\\
    Quality + Distortion & 0.920 & -- & 0.829\\
    \hline
    All Tasks (LIQE) & 0.922 & 0.894 & 0.831\\
     \bottomrule
   \end{tabular}
\end{table}

\subsection{Realignment of Quality Annotations}~\label{subsec:realign}
Despite the promising performance on each individual dataset, it is unclear whether the proposed LIQE 
suffices to
realign well the quality annotations from different IQA datasets in a common perceptual scale. In this subsection, we conduct two sessions of experiments to probe this question both qualitatively and quantitatively.

\noindent \textbf{Qualitative Results}. We rely on the gMAD~\cite{ma2020group} competition to obtain qualitative results, which seeks pairs of images that are estimated to be of similar quality by one model but of substantially different quality according to another model.
As a result, 
at least one of the two models will be falsified with 
inconsistent judgments against human opinions. We combine the full Waterloo Exploration Database~\cite{ma2017waterloo} and SPAQ~\cite{fang2020perceptual} to build a gMAD playground with both realistic and synthetic distortions. We let LIQE compete against UNIQUE~\cite{zhang2021uncertainty}, both of which are trained jointly on all six datasets stated in Sec.~\ref{subsec:exp_setup}. As shown in Fig.~\ref{fig:gMAD}, UNIQUE makes similar quality predictions for pairs of images in (a) and  (b), which  is clearly inconsistent with human opinions as well as quality predictions by LIQE. When the roles of the two models switch, UNIQUE claims that the top image in the pair of (c) is of higher quality than the bottom one. However, both images in pair (c) are of low quality to humans, which are contaminated by realistic motion blur and synthetic JPEG compression, respectively. In contrast, LIQE makes a successful defense by rating both images in pair (c) as poor perceived quality. Similar conclusions can be drawn from pair (d).

\noindent \textbf{Quantitative Results.} MOSs of images obtained in different sessions of subjective testing~\cite{sheikh2006statistical} or via different testing methodologies~\cite{perez2020pairwise} are not directly comparable. This hinders direct combination of MOSs from different datasets for training and evaluation. To test quantitatively the perceptual scale realignment performance of LIQE, we follow the method in~\cite{sheikh2006statistical}, and sample $150$ images from the six datasets, on which we conduct a separate perceptual scale realignment experiment to obtain the realigned MOSs. After that, we fit six monotonic nonlinear mapping functions  to compute the realigned MOSs for all six datasets separately. We list in Table~\ref{tab:align} the median SRCC results of LIQE and UNIQUE~\cite{zhang2021uncertainty} on the same test sets used in Sec.~\ref{subsec:main_result}. We find that LIQE achieves a higher SRCC.

In summary, we show convincingly that the proposed multitask learning scheme enabled by the vision-language correspondence improves the 
perceptual scale realignment performance on top of the pairwise learning-to-rank training strategy (used by both UNIQUE and LIQE).

\subsection{Ablation Studies}~\label{subsec:ablation}
We conduct a series of ablation studies to verify the design rationality of LIQE following the setups in Sec.~\ref{sec:exp}. We first (1)
implement a pre-trained CLIP baseline which adopts a Likert-scale of two quality levels as in~\cite{wang2022exploring}, \ie, $c \in\mathcal{C}=\{1,2\} =$ \{``bad", ``good"\}, and also (2) fine-tune it on IQA datasets. The subsequent ablations adopt the same  multitask learning scheme as LIQE, while differing in three alternative design choices: (3) freezing the language encoder $\bm{g}_{\bm{\varphi}}$ during training; (4) training with three separate textual templates with respect to quality prediction, scene classification, and distortion type identification, respectively; (5) using equal task weightings instead of the dynamic loss weighting. As a reference, we also (6) fine-tune the visual encoder $\bm{f}_{\bm{\phi}}$ using the training strategy in~\cite{zhang2021uncertainty} with a fully connected layer attached to the visual encoder. From Table~\ref{tab:ablation}, we observe that the pre-trained CLIP model is not good at BIQA, whose performance can be  improved by finetuning on IQA datasets. Performance is negatively affected when the language encoder is frozen, which may be because quality-related concepts have not been sufficiently captured during the pre-training stage of CLIP. We also find that LIQE works better with the Likert-scale of five quality levels than two. The full method performs favorably against the variant trained with three separate textual templates, verifying the effectiveness of the proposed joint probability formulation. It is desirable to also include the dynamic loss weighting scheme, which not only achieves improvements over adopting equal task weightings but also liberates us from laborious hyperparameter tuning. Finally, the superiority of LIQE over the fine-tuned visual encoder corroborates our major contribution of multitask learning via the vision-language correspondence.

\subsection{Task Relationship Analysis}~\label{subsec:relation}
We train LIQE variants on different combinations of tasks, and summarize the mean SRCC and accuracy results over six datasets in Table~\ref{tab:relation}, from which we have three observations. First, including the task of distortion type identification is beneficial for BIQA, reflecting the cooperative relationship between them. Second, as a conceptually conflicting task, bringing only the scene classification task in neither improves nor impairs BIQA; however, the best mSRCC result is achieved when LIQE is trained on all three tasks, suggesting that the proposed multitask learning scheme successfully leverages an intermediate task like distortion type identification as a bridge in soliciting contributory features.
Third, in turn, training with BIQA neither improves scene classification nor distortion type identification, indicating that the visual encoder representation optimized for BIQA moves away from the pre-trained CLIP, which are more transferable to the other two tasks.

\section{Conclusion}
\label{sec:conclusion}
We have formulated BIQA from a multitask learning perspective via the vision-language correspondence. During training, we simultaneously optimized a pair of image and language encoders on multiple IQA datasets for BIQA, scene classification, and distortion type identification jointly. We designed three fidelity losses to train the model, and employed a simple and efficient dynamic weighting scheme to automate the weighted summation of the multi-task losses.  We presented the effectiveness of the proposed LIQE, and verified the rationality of various design choices. We also showed that the learned model realigns MOSs from different datasets in a more perceptually meaningful way. We believe the proposed multitask learning perspective will shed some light on the development of next-generation BIQA models as well as computational models for other machine vision applications. 

\section*{Acknowledgments}
The authors would like to thank Chengguang Zhu for coordinating the psychophysical experiment. This work was supported in part by Shanghai Municipal Science and Technology Major Project (2021SHZDZX0102), the Fundamental Research Funds for the Central Universities, and the National Natural Science Foundation of China under Grants 61901262 and 62071407. 

{\small
\bibliographystyle{ieee_fullname}
\bibliography{Weixia}
}

\end{document}